\documentclass[letterpaper, 10 pt, conference]{ieeeconf}  

\IEEEoverridecommandlockouts                              

\usepackage{caption}
\usepackage{cite}
\usepackage{subcaption}
\usepackage{graphicx}
\usepackage{amsmath}
\usepackage{amssymb}
\usepackage{wrapfig,lipsum,booktabs}
\usepackage{algorithm}
\usepackage[ruled,vlined,algo2e]{algorithm2e}

\usepackage{theoremref} 
\usepackage{thmtools}
\usepackage{thm-restate} 
\usepackage{amsxtra} 

\usepackage{textcase}

\newcommand{\gaifo}{\textsc{gai}f\textsc{o}}
\newcommand{\gail}{\textsc{gail}}
\newcommand{\ifo}{\textsc{i}f\textsc{o}}
\newcommand{\vgaifo}{\textsc{vgai}f\textsc{o}}
\newcommand{\vgaifoso}{\textsc{vgai}f\textsc{o}-\textsc{so}}
\newcommand{\bco}{\textsc{bco}}
\newcommand{\bc}{\textsc{bc}}
\newcommand{\tcn}{\textsc{tcn}}
\newcommand{\irl}{\textsc{irl}}
\newcommand{\rl}{\textsc{rl}}
\newcommand{\ail}{\textsc{ail}}
\newcommand{\cnn}{\textsc{cnn}}
\newcommand{\ppo}{\textsc{ppo}}
\newcommand{\pilqr}{\textsc{pilqr}}
\newcommand{\gan}{\textsc{gan}}

\newcommand{\il}{\textsc{il}}
\newcommand{\lfd}{\textsc{l}f\textsc{d}}

\title{\LARGE \bf
Adversarial Imitation Learning from Video using a State Observer
}

\author{Haresh Karnan$^{1}$, Faraz Torabi$^{2}$, Garrett Warnell$^{3}$ and Peter Stone$^{4}$
	\thanks{*This work has taken place in the Learning Agents Research Group (LARG) at UT Austin. LARG research is supported in part by NSF (CPS-1739964, IIS-1724157, NRI-1925082), ONR (N00014-18-2243), FLI (RFP2-000), ARO (W911NF-19-2-0333), DARPA, Lockheed Martin, GM, and Bosch. Peter Stone serves as the Executive Director of Sony AI America and receives financial compensation for this work. The terms of this arrangement have been reviewed and approved by the University of Texas at Austin in accordance with its policy on objectivity in research.}
	\thanks{$^{1}$Haresh Karnan is with the Department of Mechanical Engineering, The University of Texas at Austin, USA
		{\tt\small haresh.miriyala@utexas.edu}}%
	\thanks{$^{2}$Faraz Torabi is with the Department of Computer Science, The University of Texas at Austin, USA
	{\tt\small faraztrb@cs.utexas.edu}}%
	\thanks{$^{3}$Garrett Warnell is with Army Research Laboratory, USA and the Department of Computer Science, The University of Texas at Austin, USA
		{\tt\small garrett.a.warnell.civ@army.mil}}%
	\thanks{$^{4}$Peter Stone is with the Department of Computer Science, The University of Texas at Austin and Sony AI, USA
		{\tt\small pstone@cs.utexas.edu}}%
}

\begin{document}

\maketitle
\thispagestyle{empty}
\pagestyle{empty}

\begin{abstract}

The imitation learning research community has recently made significant progress towards the goal of enabling artificial agents to imitate behaviors from video demonstrations alone.
However, current state-of-the-art approaches developed for this problem exhibit high sample complexity due, in part, to the high-dimensional nature of video observations.
Towards addressing this issue, we introduce here a new algorithm called Visual Generative Adversarial Imitation from Observation using a State Observer (\vgaifoso{}).
At its core, \vgaifoso{} seeks to address sample inefficiency using a novel, self-supervised \textit{state observer}, which provides estimates of lower-dimensional proprioceptive state representations from high-dimensional images.
We show experimentally in several continuous control environments that \vgaifoso{} is more sample efficient than other \ifo{} algorithms at learning from video-only demonstrations and can sometimes even achieve performance close to the Generative Adversarial Imitation from Observation (\gaifo{}) algorithm that has privileged access to the demonstrator's proprioceptive state information. 

\end{abstract}


\section{Introduction}

Imitation Learning (\il{}) \cite{schaalil, argall} is a framework in which an autonomous agent learns to imitate a demonstration provided by an expert agent, typically in the form of state and control signals. Specifically, in this work, we are interested in the problem of Imitation from Observation (\ifo{}), a sub-problem of \il{} that does not assume access to the control signals in the expert demonstrations. Several \ifo{} algorithms proposed in the past \cite{gaifo, gaifo_proprio, bco, faraz_survey, dealio, contexttranslate, ilpo, rlvid, tpil, voila} have been shown to be successful at imitating expert's state-only demonstrations in several continuous control, robotics domains.

While these prior \ifo{} algorithms have enjoyed some success, they typically assume that the demonstration includes proprioceptive state information (i.e, the most basic, internal state information that is available to the agent such as joint angles and velocities of a legged robot). While some algorithms, such as \gaifo{}, can learn from video demonstrations instead, the additional sample complexity incurred as a result can be detrimental in domains such as robotics where data collection during online learning is costly. 
For example, for the common benchmark task of Hopper-v2 on MuJoCo \cite{mujoco}, we found that an \ifo{} algorithm using video demonstrations took over three times more timesteps to learn a good policy, compared to the case where a proprioceptive demonstration was available (see Fig. \ref{fig:gaifo_vgaifo_diff}). This problem is especially relevant in scenarios where proprioceptive information about the demonstrator might not be available at all. For example, it may be expensive or impossible to fit specialized sensors on the demonstrator agents to record the proprioceptive state information. In some cases, we may not have access to the demonstrator at all---for example, when learning to imitate skills from YouTube videos, or learning using video demonstrations collected from robots with proprietary code that one would like to mimic on the same or different hardware.

\begin{figure*}
	\centering
	\includegraphics[scale=0.23]{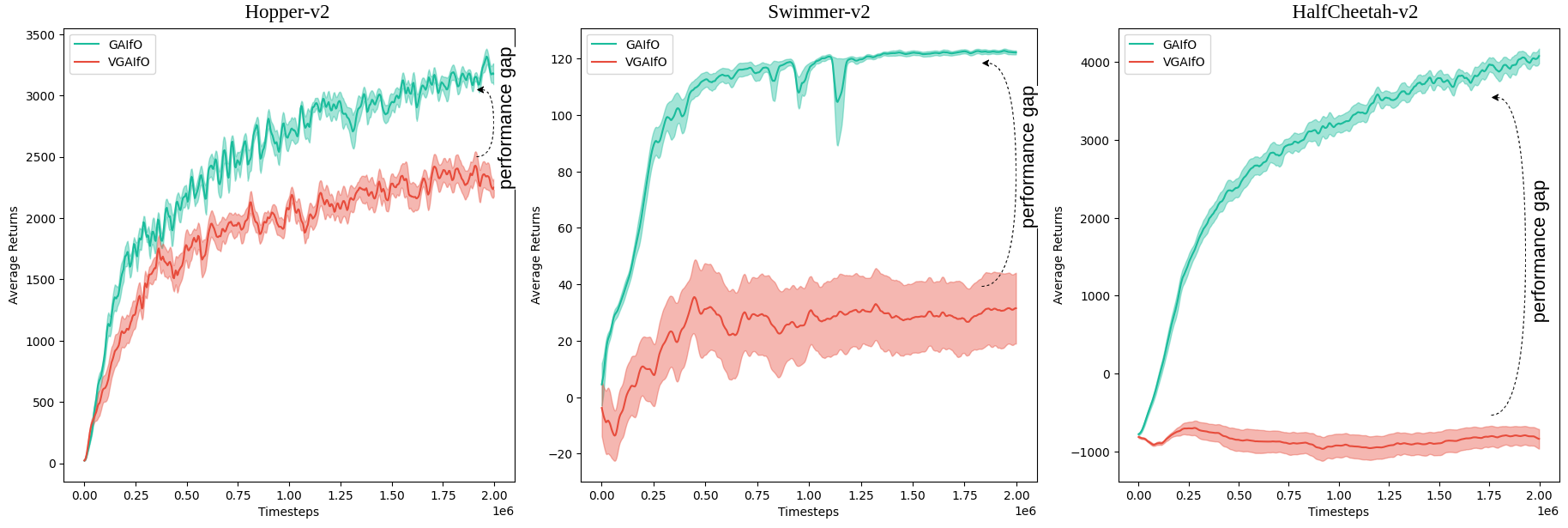}
	\caption{
		Demonstrating the need for more sample efficiency and improved performance in imitation learning from video-only demonstrations. Here we show the learning curves of two representative \ail{} algorithms, \gaifo{}---that learns to imitate an expert with privileged access to proprioceptive-state-only demonstrations and \vgaifo{}---that learns to imitate an expert from video-only demonstrations, without access to expert's proprioceptive state information. The x-axis shows number of timesteps of interactions for the imitator agents with the environment and the y-axis shows the task reward function from OpenAI gym \cite{openai} used only for evaluation. Evident from the performance gap in their respective learning curves, we notice that \gaifo{} is more sample efficient than \vgaifo{} in imitation learning. In this work, we seek to improve both sample efficiency and imitation learning performance by proposing a new algorithm called \vgaifoso{}. 
	}
	\label{fig:gaifo_vgaifo_diff}
\end{figure*}

To alleviate this sample inefficiency problem, an extension to \gaifo{} for learning from video-only demonstrations was proposed \cite{gaifo_proprio} (referred to as \vgaifo{} hereafter) that was shown to have better sample efficiency compared to \gaifo{}. 
\vgaifo{} achieves improved sample efficiency when learning to imitate video-only demonstrations by leveraging the available proprioceptive state information from the imitator. Although \vgaifo{} is better than \gaifo{} at imitation from video-only demonstration data, there is still a significant gap in performance compared to \gaifo{} that can also imitate from proprioceptive state-only expert demonstrations. Fig.\ref{fig:gaifo_vgaifo_diff} shows this difference in performance between \gaifo{} (which has privileged access to proprioceptive states of the expert) and \vgaifo{} (which has access only to visual observations of the expert). The performance gap between these two algorithms motivates us here to seek a new method to improve both sample efficiency and performance when learning to imitate from video-only demonstrations.

We hypothesize that improved sample efficiency when learning to imitate from video-only demonstrations can be achieved by ensuring that the imitation learning process itself happens over lower-dimensional quantities like proprioceptive state information, rather than high-dimensional visual observations for both the generator and the discriminator. Therefore, we propose here a novel self-supervised \textit{state observer} function that estimates the proprioceptive states of the agent from high-dimensional observations. 
Based on this state observer, we propose a novel \ifo{} algorithm called Visual Generative Adversarial Imitation from Observation using a State Observer (\vgaifoso{}). 
In \vgaifoso{}, the state observer is jointly learned along with the generator (imitator's policy) and the discriminator within an adversarial \ifo{} framework \cite{gaifo}. We experimentally show that using the low-dimensional proprioceptive state predictions of the state observer as input to the discriminator network leads to improved sample efficiency and performance when learning to imitate from video-only demonstrations. Although the use of \ail{} for \ifo{} is not a new concept, we show here for the first time that using a state observer to estimate proprioceptive states from video demonstrations significantly improves sample efficiency when learning to imitate from video-only observations.

This paper makes three main contributions: \textit{1)} we show that there exists a significant gap in sample efficiency between the two \ifo{} algorithms \gaifo{} (which has privileged access to proprioceptive states of the expert) and \vgaifo{} (which \textit{does not} have access to proprioceptive states of the expert); \textit{2)} we propose a novel algorithm called \vgaifoso{}, which explicitly seeks to perform imitation learning over low-dimensional quantities such as proprioceptive states of the agent, using a novel state observer network; \textit{3)} we show on a suite of MuJoCo benchmark environments \cite{mujoco, openai} that \vgaifoso{} narrows the gap in performance between \gaifo{} and \vgaifo{} by improving the sample efficiency, and performs better than other baseline \ifo{} algorithms.


\section{Related Work}
\label{sec:relatedwork}


\textbf{Learning from Demonstration.} Learning from Demonstration (\lfd{}) is a machine learning framework in which, an autonomous agent learns to imitate an expert demonstration of a behavior. The demonstrations usually contains sequences of state-action pairs 
generated by an expert policy when deployed in the environment. Several algorithms have been proposed to solve the imitation learning problem, and they can be broadly classified into two main categories \cite{gaifo_proprio}---Behavior Cloning (\bc{}) \cite{bainandsommut, ross2011, daftry2016} in which the imitative policy is learned directly from demonstrations using supervised learning, and Inverse Reinforcement Learning (\irl{}) \cite{apprenticeship, bagnell, bakerAU} where a reward function is first learned from the demonstrations, which is then used to learn the policy using Reinforcement Learning (\rl{}) \cite{suttonandbarto}.
Recently, a third category of imitation learning algorithms, called Adversarial Imitation Learning (\ail{}) was proposed \cite{gail, valuedice, pwil}, which utilizes an adversarial learning setup similar to a \gan{} \cite{gan} that learns both a discriminator network that classifies experiences from the expert and the imitator, and a generator network (imitator's policy) that imitates the expert demonstrations.  Generative Adversarial Imitation Learning (\gail{}) \cite{gail} was one of the earliest proposed \ail{} algorithms that was shown to be very successful at learning to imitate expert demonstrations. One shortcoming is that all of these algorithms, including \gail{}, assumes access to actions performed by the expert in the demonstration, which may not be available necessarily.

\begin{figure*}
	\centering
	\includegraphics[width=\linewidth]{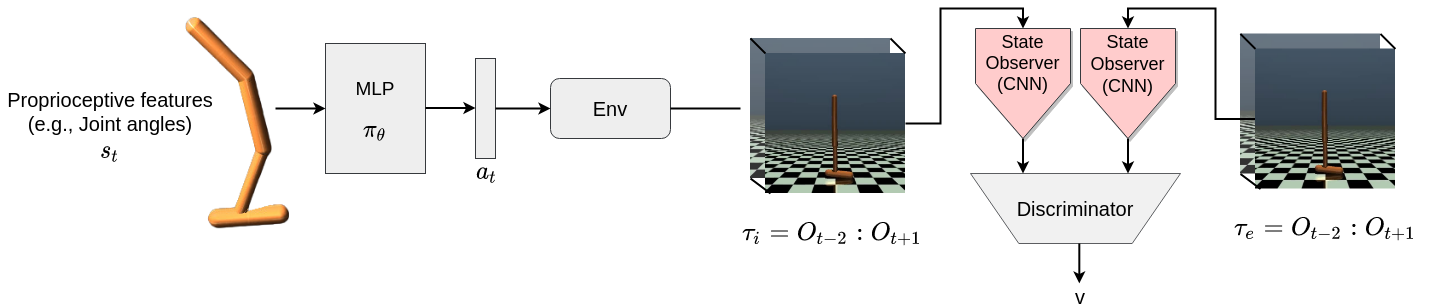}
	\caption{
		A diagrammatic representation of the \vgaifoso{} algorithm proposed in this work.
		Different from existing \ail{} methods, we propose here to utilize a novel \textit{state observer} module, which serves to simplify discriminator training. More specifically, the state observer learns to map high-dimensional visual observations to low-dimensional proprioceptive states of the agent. While traditional methods like \vgaifo{} directly utilize the high-dimensional observations in optimizing the discriminator's objective, in this work, we instead use the low-dimensional proprioceptive state predictions of the state observer to optimize the discriminator objective. 
	}
	\label{fig:gaifoso}
\end{figure*}

\textbf{Imitation Learning from Observation-only Demonstrations.}
Fortunately, advances in the Imitation from Observation (\ifo{}) community have addressed the problem of performing imitation learning when the demonstration data lacks action information. In the \ifo{} problem setting, the demonstrations consists of states-only or observations-only sequences. Behavior Cloning from Observation (\bco{}) \cite{bco} is an \ifo{} algorithm that uses behavior cloning \cite{behaviorcloning} to learn the imitative policy given the expert's state-only demonstrations. However, behavior cloning has been shown to suffer from the well-known compounding-error issue \cite{ross2011, ross2010} and \bco{} \cite{bco} is no exception \cite{gaifo}.  Generative Adversarial Imitation from Observation (\gaifo{}) \cite{gaifo}, on the other hand, overcomes this issue by incorporating reinforcement learning and exploration, which has led to its state-of-the-art \ifo{} performance.
An extension of \gaifo{} \cite{gaifo} called \vgaifo{} \cite{gaifo_proprio} was recently proposed that improves upon the poor sample efficiency and performance of \gaifo{} when learning to imitate from video-only demonstrations of the expert. \vgaifo{} leverages the already available proprioceptive states of the imitator as inputs to the imitative policy, which makes it more sample efficient than \gaifo{} with video-only demonstrations. 
Time Contrastive Networks (\tcn{}) \cite{tcn} is another \ifo{} algorithm that was shown to be successful at imitation learning from video-only demonstrations.
\tcn{} achieves self-supervised imitation learning by first learning an embedding of the visual observations in the demonstrations using time-contrastive metric learning which is then used to drive a reinforcement learning procedure. Note that the version of \tcn{} we use as a comparison point to \vgaifoso{} is also called single-view \tcn{}. The multi-view \tcn{} approach for \ifo{} deals with addressing the problem of viewpoint and domain mismatch, similar to other related approaches such as \textsc{tpil} \cite{tpil}, among other methods \cite{rlvid, ilpo}. 

\textbf{Sample efficiency in IfO.} Sample efficiency is a desirable property of algorithms that enable robots to acquire skills through imitation. 
Using off policy learning within the \ail{} framework has been proposed as a way to improve sample efficiency \cite{opolo, valuedice, saifo}. Another way to improve sample efficiency is to use model-based reinforcement learning techniques within \ail{} \cite{dealio}. We note that using off-policy \rl{} or model-based \rl{} to improve sample efficiency in imitation learning is orthogonal to this work. However, the techniques introduced in this work are not restricted to on-policy \il{} and can be combined with off-policy and model-based \rl{} algorithms as well to further improve performance. We view this work as a stepping stone towards the greater goal of \ifo{} by focusing on the problem of sample inefficiency when learning to imitate from video-only observations, a relevant problem in robotics domains. While addressing domain, viewpoint, and embodiment mismatches are also important challenges to address in imitation learning, we leave incorporating them into \vgaifoso{} for future work.



\begin{algorithm}
	\caption{\vgaifoso{}}
	\label{alg}
	\SetAlgoLined
	\KwIn{Initialize imitator policy $\pi_\theta$ randomly}\;
	Initialize discriminator network $D_\phi$ randomly\\
	Initialize state observer network $S_\eta$ randomly\\
	Obtain video-only demonstrations $\tau_e=\{\tau_{e_1}, \tau_{e_2}, \ldots\}$, 
	where $\tau_{e_k}=\{O_1, O_2, \ldots\};$ \\
	\For{$i=0,1,2, \ldots N$}{
		Execute $\pi_\theta$ and obtain paired proprioceptive-and-visual state trajectories $\tau_i=\{\tau_{i_1}, \tau_{i_2}, \ldots\}$\;\\ where $\tau_{ij}=\{(s_1, O_1), (s_2, O_2), \ldots\};$ \\ 
		
		$\textbf{Train State Observer : }$Update parameters $\eta$ of $S_{\eta}$ using the regression loss $-\mathbb{E}_{\tau_{i}}[(S_\eta(O_t) - s_t)^2]$\ for $N_s$ epochs and freeze network parameters $\eta$; \\
		
		Update parameters $\phi$ of $D_{\phi}$ using gradient descent to minimize \\ $-\left(\mathbb{E}_{\tau_{i}}[\log (D_{\phi}(\hat{s}, \hat{s'}))] + \mathbb{E}_{\tau_{e}}[ \log(1 - D_{\phi}(\hat{s}, \hat{s'}))\right)$\; \\ where $\hat{s} = S_\eta(O)$, $\hat{s'} = S_\eta(O')$; \\
		
		Update parameters $\theta$ of $\pi_\theta$ using \textit{PPO} updates with reward $-[\log D_{\phi}(\hat{s}, \hat{s'})]$;
	}
\end{algorithm}

\begin{figure*}[!t]
	\centering
	\includegraphics[scale=0.4]{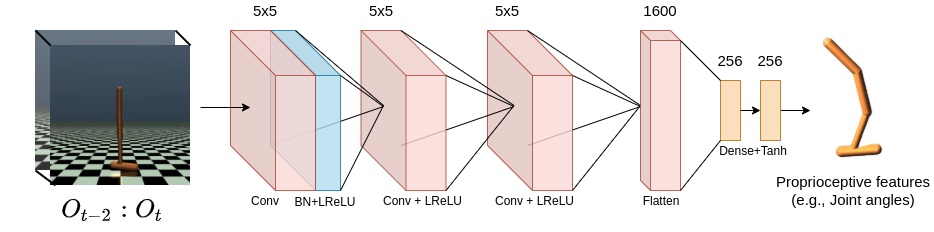}
	\caption{Network architecture of the state observer function proposed in this work. The state observer takes as input three consecutive visual frames of observations, and produces as output a prediction $\hat{s}_t$ of the proprioceptive features of the agent.}
	\label{fig:stateobserver}
\end{figure*}

\section{Method}
\label{sec:background}
In this section, we begin by formally introducing the imitation learning problem. We then describe the algorithm introduced in this work--- Visual Generative Adversarial Imitation from Observation using a State Observer (\vgaifoso{}). 

\subsection{Preliminaries}
We formulate sequential decision making as a Markov Decision Process (MDPs) \cite{suttonandbarto}. An MDP is a tuple $\langle \mathcal{S}, \mathcal{A}, \mathcal{R}, \mathcal{T}, \gamma, \rho_0 \rangle$ where $\mathcal{S}$ and $\mathcal{A}$ are state and action spaces of the agent respectively, and $\mathcal{R}$ is a task reward function. When the agent performs an action $a_t \in \mathcal{A}$ in the environment at a state $s_t \in \mathcal{S}$, it transitions to a new state $s_{t+1} \in \mathcal{S}$ according to the transition dynamics $\mathcal{T}(s_{t+1} | s_t, a_t)$ that is unknown to the agent. $\gamma$ is the discount factor that controls the relative long term utility of the reward and $\rho_0$ is the initial state distribution of the environment. Since we are concerned with visual observations, we denote the observation space of the agent as $\mathcal{O}$. We are interested in learning a policy $\pi : \mathcal{S} \rightarrow \mathcal{A}$ that the agent can use to select actions that result in behavior similar to what was demonstrated. In the imitation learning setting, agents do not receive a reward $r_t \in \mathcal{R}$ from the environment. Instead, the imitative agents have access to an expert demonstration $\mathcal{D}_E=\{(s_0, a_0), (s_1, a_1) \ldots\}$ consisting of state-action pairs. However, in this work, we focus on the problem of Imitation from Observation (\ifo{}) consisting of video-only demonstrations $\mathcal{D}_E=\{O_0, O_1, \ldots\}$ where $O \in \mathcal{O}$.

The camera observations of the robots considered in this work are floating third person views of the robot performing the task in the environment, rendered using the MuJoCo simulator \cite{mujoco, openai}, as shown in Fig. \ref{fig:gaifoso}. 

\begin{figure*}
	\centering
	\includegraphics[scale=0.3]{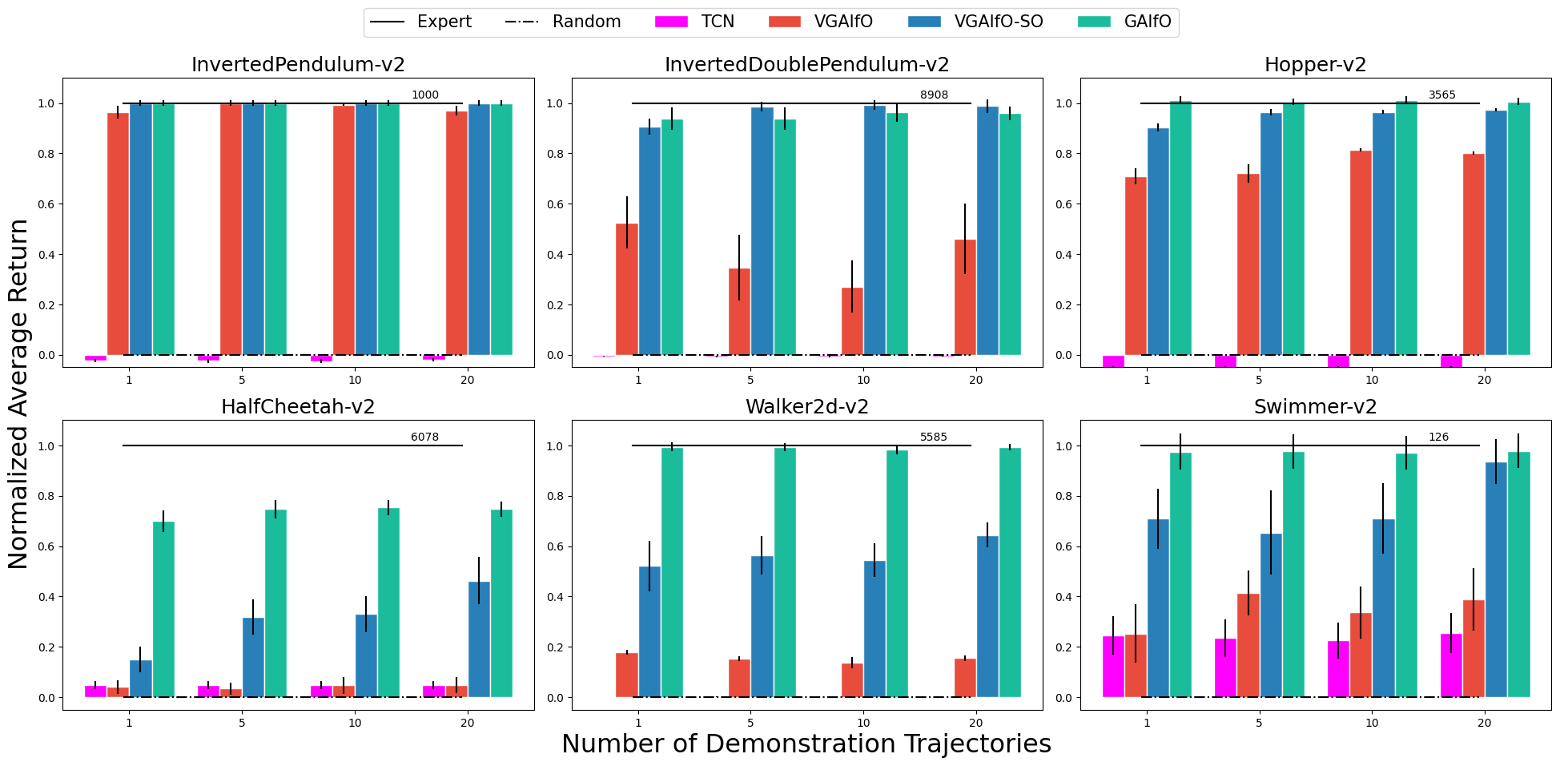}
	\caption{Performance of different \ifo{} algorithms with varied numbers of expert demonstration trajectories, trained for two million timesteps of interaction with the environment. We see that \gaifo{} \cite{gaifo} performs the best due to its privileged access to expert's proprioceptive state information. \vgaifoso{}, the algorithm introduced in this work, outperforms \vgaifo{} \cite{gaifo_proprio} and \tcn{} \cite{tcn} on all six environments and achieves performance similar to \gaifo{} in \textit{InvertedPendulum}, \textit{InvertedDoublePendulum} and \textit{Hopper}, without access to the expert's true proprioceptive states.}
	\label{fig:barplots}
\end{figure*}

\subsection{Visual Generative Adversarial Imitation from Observation using a State Observer}

We adopt the general adversarial framework for \ifo{} as initially proposed in \gaifo{} \cite{gaifo}, and also followed in \vgaifo{} \cite{gaifo_proprio}. In this \ifo{} framework, a parameterized generator function (imitative policy) $\pi_\theta$ represented by a Multi-Layer Perceptron (MLP) takes as its inputs proprioceptive states of the agent and acts in the environment, so as to produce state transitions that imitate the expert demonstrations. Observations of the imitator's behavior are then used to train the parameterized discriminator function $D_\phi$ to classify between observation transitions from the imitator and the demonstration sequence. The output of the discriminator network is then used as a reward signal to drive an update to the imitator via \rl{}. Both the discriminator and the generator are updated using the adversarial \ifo{} objective proposed in \gaifo{} \cite{gaifo}. We hypothesize that the poor sample efficiency of \vgaifo{} compared to \gaifo{}, as shown in Fig. \ref{fig:gaifo_vgaifo_diff} is due to the fact that the discriminator objective is optimized over high-dimensional images. 

To solve the sample efficiency problem when learning from video-only demonstrations, we propose the novel \vgaifoso{} algorithm. In \vgaifoso{}, a state observer function represented as a convolutional neural network (\cnn{}) (see Fig.\ref{fig:stateobserver}) is learned to circumvent optimizing the discriminator objective directly from high-dimensional visual observations. The parameterized state observer function $\mathcal{S}_\eta$ is trained using self-supervision. More specifically, $\mathcal{S}_\eta$ is trained to minimize the mean-squared error $\mathbb{E}_{\tau_{i}}[(S_\eta(O_t) - s_t)^2]$ between the predicted and ground truth proprioceptive states of the agent. The training data is obtained from already available experience gathered by the imitator when exploring in the environment. In our implementation, we alleviate partial observability issues by performing frame stacking with three consecutive frames of visual observations as is commonly done in the reinforcement learning community \cite{dqn}, although one can also use other approaches such as \textsc{lstm} \cite{lstm} for sequence modelling. 

Note that in this work, we neither assume access to the proprioceptive states of the expert nor its actions in the demonstration. We posit that a more natural, and less restrictive form of the \il{} problem is to imitate from video-only demonstrations, assuming we have access to the proprioceptive state and visual observations of the imitator that we have physical access to. 
The \vgaifoso{} algorithm is provided in Alg. \ref{alg}. Note that in Alg. \ref{alg}, $O'$ and $s'$ are the sequentially next observation  and states after $O$ and $s$ respectively.
When training the discriminator, the state observer network is frozen and used to infer proprioceptive states from high-dimensional visual observations of both the expert and the imitator. Although one can directly use the known proprioceptive states $s_t$ in place of predicted proprioceptive states $\hat{s}_t$ for the imitator agent, we find empirically that using predicted proprioceptive states performs better. 
The discriminator and generator networks are updated with the well-known \gan{}-like loss for \ifo{} \cite{gaifo}. The imitator policy is updated using the \ppo{} \cite{ppo, ppoilya} algorithm. All three networks are updated iteratively until the imitator successfully imitates the expert. 
\begin{figure}[!h]
	\centering
	\includegraphics[scale=0.30]{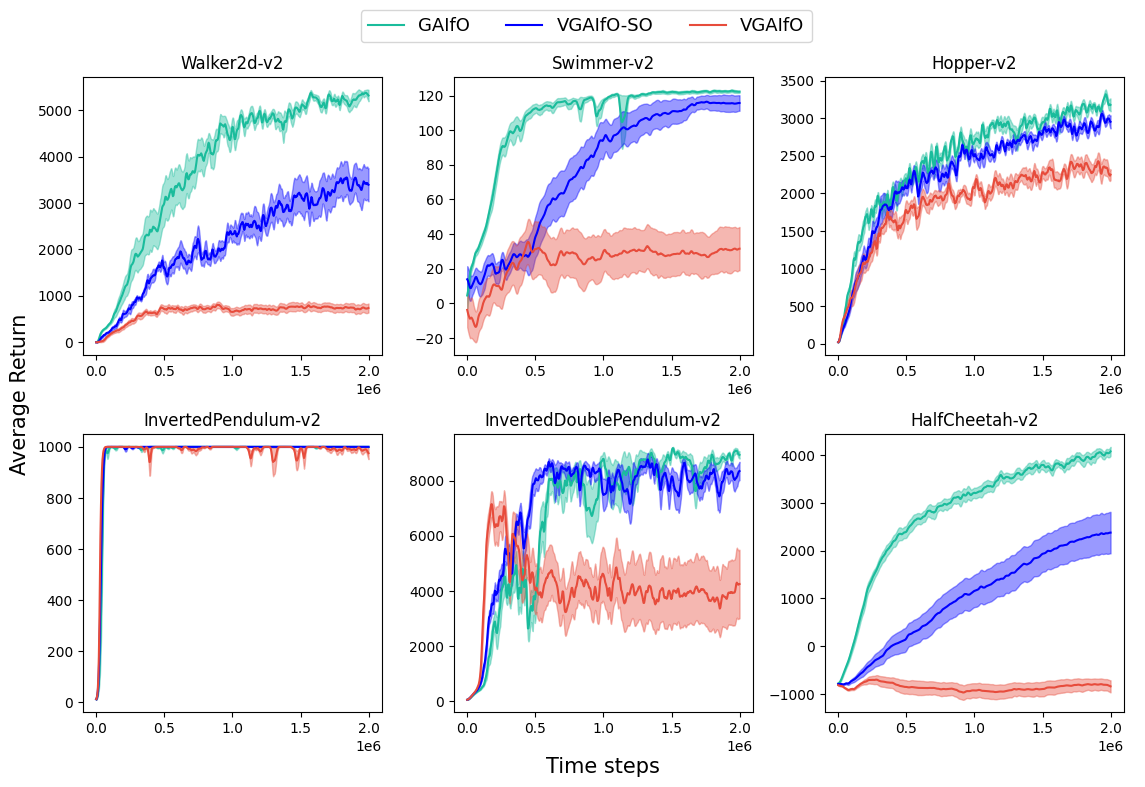}
	\caption{Learning curves of the three adversarial \ifo{} algorithms \gaifo{}, \vgaifo{} and \vgaifoso{} (ours). The learning curves here show that \vgaifoso{} has better sample efficiency than \vgaifo{} in imitation learning from video-only demonstrations and achieves performance close to the state-of-the-art \gaifo{} algorithm that has privileged access to expert's proprioceptive states. The x-axis shows timesteps of interactions with the environment and the y-axis shows the task reward from OpenAI gym \cite{openai} (used only for evaluation). We use average return as the metric to evaluate performance.}
	\label{fig:learningcurves}
\end{figure}


\section{Experiments}
\label{sec:experiments}

In this section, we perform experiments to evaluate our Visual Generative Adversarial Imitation from Observation using a State Observer (\vgaifoso{}) algorithm. Note that \vgaifoso{} is applicable in problems where the proprioceptive state representation of the agent contains all necessary information to solve the task. In manipulation domain, for example, if the task involves manipulating objects, the state representation should contain sensed poses of those objects, along with the proprioceptive state information of the manipulator. However, in this work, we restrict our analysis to the locomotion domain that we are primarily interested in, and not the manipulation domain. We hypothesize that, compared to baseline approaches, using the predictions of the state observer in optimizing the discriminator's objective improves the imitation learning performance and sample efficiency under two million environment interaction timesteps. 

\subsection{Methodology}

We test our hypothesis by evaluating \vgaifoso{} on a suite of continuous control environments in MuJoCo \cite{mujoco, openai}, that were also used to evaluate other related \il{} algorithms \cite{gail, gaifo, gaifo_proprio} in the past. We compare \vgaifoso{} with two representative \ail{} algorithms \gaifo{} (with privileged access to expert's proprioceptive states) and \vgaifo{}. Note that \gaifo{} is expected to perform better than \vgaifoso{} since it learns to imitate from low-dimensional proprioceptive state-only demonstrations of the expert. We also compare \vgaifoso{} against single-view \tcn{} \cite{tcn}, an algorithm for self-supervised imitation learning from video-only demonstrations. 
In our implementation of \tcn{}, instead of using \pilqr{} \cite{pilqr}, we use \ppo{} \cite{ppo, ppoilya} to learn the imitative policy. Note that \tcn{} also uses the proprioceptive states of the imitator as input to the imitative policy, similar to \gaifo{}, \vgaifo{}, and \vgaifoso{}, making it a fair baseline for comparison. We additionally perform analysis experiments on \vgaifoso{} to qualitatively evaluate the proprioceptive state predictions from the state observer. In all experiments reported, the results are averaged across ten different random initial seeds. The expert demonstrations in each environment are generated by a near-optimal policy trained to maximize the cumulative sum of returns. To compare the different imitation learning algorithms, we compute the Normalized Average Return metric by normalizing the returns (ground truth reward provided by OpenAI gym \cite{openai}) achieved in an environment between 0 (random policy) and 1 (demonstrator).

\subsection{Results and Discussion}

Figures \ref{fig:barplots} and \ref{fig:learningcurves} depict our experiments comparing \vgaifoso{} with \gaifo{}, \vgaifo{} and \tcn{} on the six continuous control environments in MuJoCo \cite{mujoco, openai}. We see that \vgaifoso{} performs significantly better than \vgaifo{} and \tcn{}, and achieves performance close to \gaifo{} and the expert policy in \textit{InvertedPendulum}, \textit{InvertedDoublePendulum} and \textit{Hopper}, possibly due to the state observer providing as much information to the discriminator as the raw proprioceptive states. Fig. \ref{fig:barplots} shows that \tcn{} does not achieve performance anywhere near the expert. A similar observation was made earlier by Torabi et al. \cite{gaifo_proprio}, perhaps due to the tasks considered here being cyclical in nature and not well suited to the time-dependent embeddings learned by \tcn{}. The learning curves of the three \ail{} algorithms is shown in Fig. \ref{fig:learningcurves}, validating the superior sample efficiency of \vgaifoso{} compared to \vgaifo{}.

On \vgaifoso{}, we performed a coarse hyperparameter search on number of training epochs on the three networks per adversarial training iteration and found that performing a single epoch per iteration on the networks worked best, except environments \textit{HalfCheetah, Walker2d and Swimmer} that required 20 state observer epochs per iteration. We performed similar hyperparameter searches across other baseline algorithms \gaifo{}, \vgaifo{} and \tcn{} and report results for the best set of hyperparameters.

\textbf{State Observer Analysis.} While the state observer is trained exclusively using experience from the imitator, we use it to predict proprioceptive states for both the imitator and the demonstrator. Therefore, it is important to ensure that it behaves as expected on the demonstration observations as well. To determine if the state observer is actually predicting the correct proprioceptive state, we study the prediction errors of the state observer in estimating the expert's proprioceptive states averaged across the demonstration sequence. Fig. \ref{fig:so_learning_curve} shows the average $L^2$ norm of the prediction error after every state observer update epoch. While the prediction errors are initially high, these errors decrease with more and more training. Therefore, we conclude that the state observer is also learning to predict the true proprioceptive state of the expert agent as well.

\begin{figure}
	\centering
	\includegraphics[scale=0.33]{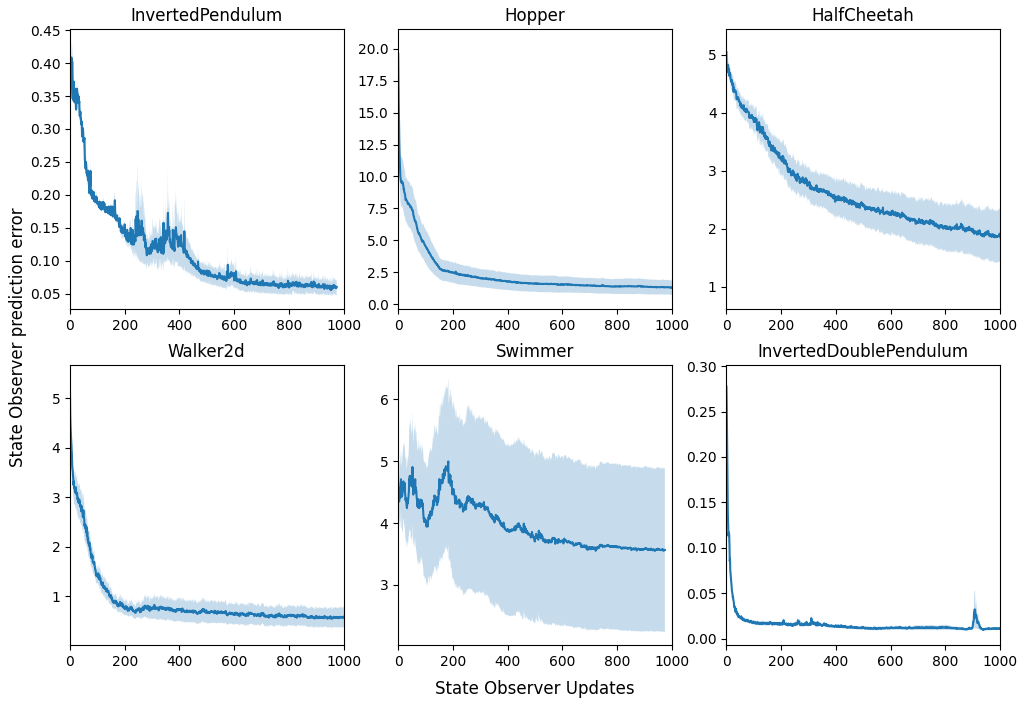}
	\caption{\textbf{State Observer Analysis} Average $L^2$ norm of proprioceptive state prediction error of the state observer on the expert's demonstration sequence, per update iteration. We see that in all environments, the error has a downward trend which shows that as learning progresses, the state observer tends towards predicting the true proprioceptive states of the demonstrator (unseen data) as well as the imitator (training data). }
	\label{fig:so_learning_curve}
\end{figure}


\section{Conclusion}
\label{sec:conclusion}
\textbf{Summary.} 
In this work, we introduced Visual Generative Adversarial Imitation from Observation using a State Observer (\vgaifoso{}), an \ifo{} algorithm that learns to imitate an expert policy from video-only demonstrations. We showed that, by regressing an intermediate state representation such as the proprioceptive states of the agent from visual observations using a \textit{state observer network}, both the imitation learning performance and sample efficiency can be improved significantly. We compared our approach against similar baselines and showed that \vgaifoso{} learns to imitate the expert with better sample efficiency, and also achieves better overall imitation performance.

\textbf{Limitations and Future Work.} While we have shown that \vgaifoso{} performs as well as \gaifo{} in some domains, it does not in all experimental domains we studied (e.g., in some relatively harder tasks such as \textit{HalfCheetah} and \textit{Walker2d}, there still exists a gap in performance between \gaifo{} and \vgaifoso{}). This performance gap can potentially be narrowed further by extending \vgaifoso{} with recent advances in the reinforcement learning community involving image augmentations towards sample efficient learning \cite{rad, ilyaiclr21}. Another approach is to jointly train the state observer and the discriminator with their combined losses, end-to-end, to further improve performance. In this work, we assume no viewpoint or domain mismatch between the demonstrator and the imitator since the focus of this work is in improving sample efficiency when learning to imitate from video-only demonstrations. However, viewpoint and domain mismatch are also important challenges to overcome when performing imitation learning from video-only demonstrations in the real-world. \vgaifoso{} can be potentially combined with algorithms such as \textsc{tpil} \cite{tpil} to overcome this limitation.

\addtolength{\textheight}{-8cm}   

\bibliographystyle{IEEEtran}
\bibliography{IEEEabrv,IEEEexample}

\end{document}